%% file: main_jm_final.tex
\documentclass{article}

\usepackage{iclr2027_conference,times}
\input{math_commands.tex}

\usepackage{adjustbox}
\usepackage{amsmath}
\usepackage{amssymb}
\usepackage{booktabs}
\usepackage{colortbl}
\usepackage{float}
\usepackage{flafter}
\usepackage{graphicx}
\usepackage{multirow}
\usepackage{placeins}
\usepackage{subcaption}
\usepackage{tikz}
\usepackage{url}
\usepackage{xcolor}
\usepackage{hyperref}

\usepackage{multirow}
\usepackage{adjustbox}
\usepackage[table]{xcolor}

\definecolor{lightgray}{RGB}{245,245,245}

\definecolor{citeteal}{RGB}{75,120,55}
\hypersetup{
    colorlinks,
    linkcolor=blue,
    citecolor=citeteal,
    urlcolor=citeteal
}

\newcommand{\captionheading}[1]{\textbf{#1.}}


\definecolor{HeaderNavy}{RGB}{28,44,84}
\definecolor{AvgHead}{RGB}{88,94,104}
\definecolor{AvgCell}{RGB}{232,233,236}
\definecolor{Group3B}{RGB}{201,219,247}
\definecolor{Band3B}{RGB}{236,242,253}
\definecolor{Band3Bours}{RGB}{221,232,250}
\definecolor{Group7B}{RGB}{222,214,241}
\definecolor{Band7B}{RGB}{243,240,251}
\definecolor{Band7Bours}{RGB}{231,224,247}
\definecolor{GainGreen}{RGB}{0,128,58}
\definecolor{DropRed}{RGB}{197,42,42}

\newcommand{\gain}[1]{\textcolor{GainGreen}{\footnotesize$\blacktriangle$\,#1}}
\newcommand{\drop}[1]{\textcolor{DropRed}{\footnotesize$\blacktriangledown$\,#1}}
\newcommand{\best}[1]{\textbf{#1}}
\newcommand{\avg}[1]{\cellcolor{AvgCell}#1}

\usepackage{float}

\title{Architectural Sampling: Test-Time Scaling via Computational Diversity in Frozen Vision-Language Models}

\author{%
  Akshit Singh\textsuperscript{1} \quad
  Shyam Marjit\textsuperscript{2} \quad
  Wei Lin\textsuperscript{3} \quad
  Leonid Karlinsky\textsuperscript{3} \quad
  M.\ Jehanzeb Mirza\textsuperscript{3} \\[0.5em]
  \textsuperscript{1}Dartmouth College \quad
  \textsuperscript{2}IISc Bangalore \quad
  \textsuperscript{3}Xero \\[0.8em]
  \centering
  Project Page: \href{https://akshit21112002.github.io/Architectural-Sampling/}%
                     {https://akshit21112002.github.io/Architectural-Sampling/}
}

\iclrfinalcopy

\begin{document}

\maketitle

\begin{abstract}
Test-time scaling often seeks better answers by sampling multiple responses from a frozen model, yet conventional temperature sampling generates every candidate along the same fixed computation path. We introduce architectural sampling, a training-free method that generates candidates through distinct forward computations by reusing selected blocks of decoder layers. Varying the block location and repetition count introduces computational diversity without updating model weights or adding auxiliary parameters. 
Across five Qwen checkpoints and twelve multimodal benchmarks, architectural sampling improves $\mathrm{pass}@9$ over standard-path temperature sampling
by 6.58 percentage points on average at the same nine-candidate budget. 
Reusing early layers yields the strongest gains, and the improvement in candidate coverage persists even under greedy decoding. 
The resulting candidates show lower lexical overlap and improve accuracy when used as rollouts for label-free test-time reinforcement learning. These findings extend the benefits of our architectural sampling beyond candidate coverage, demonstrating more effective learning from a model's own outputs. 

\end{abstract}
\begin{flushright}
\begin{minipage}{0.72\linewidth}
\raggedleft\small\itshape
``No man ever steps in the same river twice, for it is not the same river and
he is not the same man.''
\par\textup{--- Heraclitus}
\end{minipage}
\end{flushright}
\begin{figure}[H]
    \centering
    \includegraphics[pagebox=cropbox,width=\linewidth]{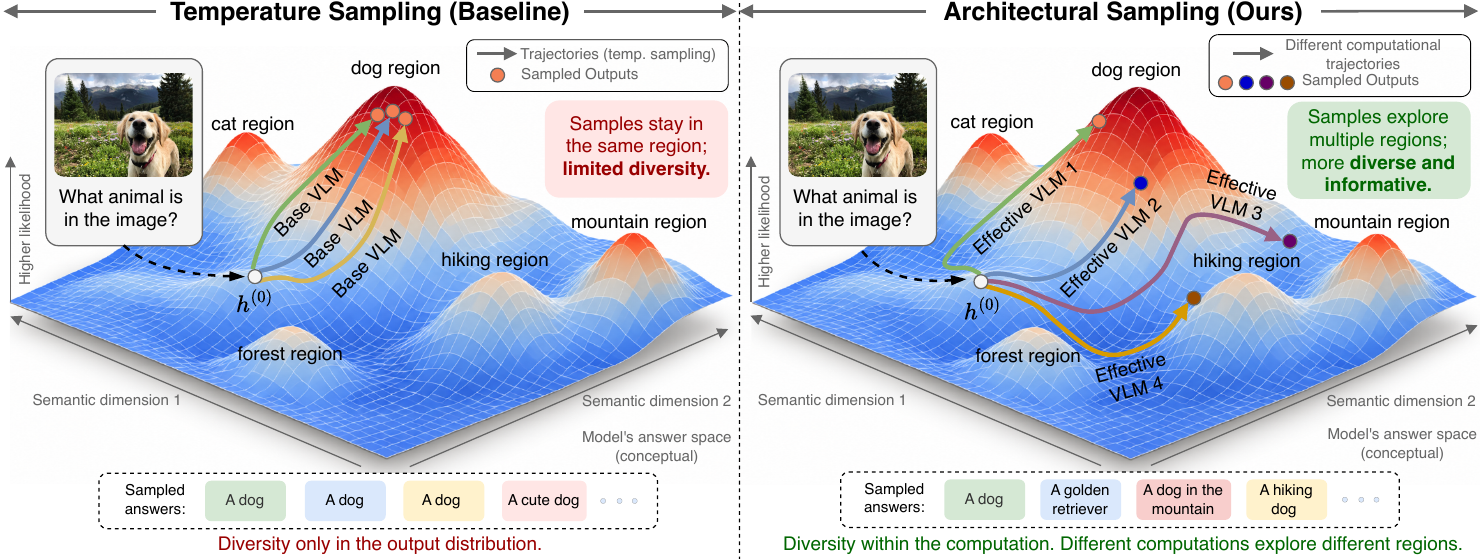}
    \caption{\captionheading{Two Sources of Candidate Diversity} Temperature sampling (left) draws several outputs from the same standard computation path, starting from the initial multimodal representation $h^{(0)}$. Architectural sampling (right) changes the computation path from the same $h^{(0)}$ before producing one candidate from each path. The background is a conceptual illustration of the answer space.}
    
    \label{fig:motivation}
\end{figure}

\begin{figure}[!t]
\centering
\includegraphics[width=1.0\linewidth]{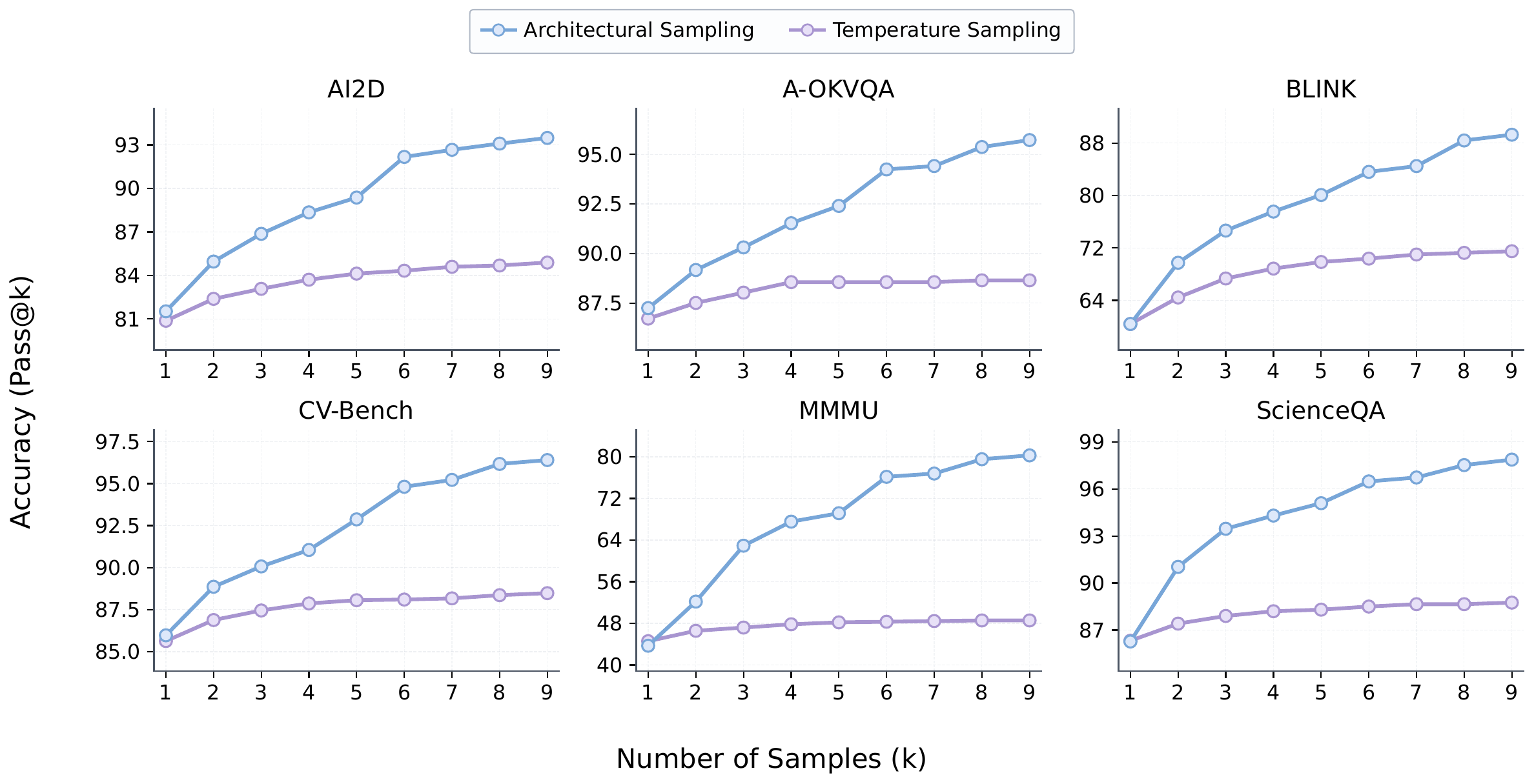}
\caption{\captionheading{Architectural sampling scales more favorably with candidate budget} $\mathrm{pass}@k$ of Qwen3-VL-8B across six multimodal benchmarks as the candidate budget grows. Architectural sampling continues to improve candidate coverage as the pool grows, while temperature sampling plateaus earlier.}
\label{fig:qwen3_8b_scaling}
\end{figure}

\input{intro_try}

\section{Related Work}

Architectural sampling sits between test-time scaling and recursive
computation. The former spends additional inference-time computation to obtain
better outputs; the latter changes the computation performed within the model.
We review these two lines in turn.

\paragraph{VLMs and Test-Time Scaling.}
Test-time scaling seeks more from a fixed checkpoint through repeated
generation, aggregation, search, or online adaptation
\citep{snell2025scaling,kaya2026efficient,singh2026ttrv}.

These methods use additional responses in different ways. Self-consistency
aggregates independently sampled reasoning paths \citep{wang2022self}. Visual
self-ensembling aggregates answers to paraphrases of the same question
\citep{nguyen2024improving}. VisuoThink and VReST explore reasoning paths with
tree search \citep{wang2025visuothink,zhang2025vrest}. Test-time augmentation
varies the input and aggregates token-level predictions
\citep{kaya2026efficient}, while TTRV turns response frequency and the entropy
of the empirical output distribution into rewards for online adaptation
\citep{singh2026ttrv}. These methods differ in how they use the responses, but
each begins with the alternatives the model has produced.

Temperature and nucleus sampling are common ways to build such candidate pools
\citep{holtzman2019curious}. They vary token selection, but every response still
traverses the same checkpoint and ordered sequence of layers. Candidate
coverage is commonly measured with $\mathrm{pass}@k$, the fraction of examples
for which at least one of $k$ responses is correct \citep{chen2021evaluating}.
 Architectural sampling introduces a complementary source of diversity by varying which layer windows are reused and how often. It remains compatible with stochastic decoding, allowing computational and token-level variation to be combined. 

\paragraph{Recursive Architectures and Inference-Time Layer Reuse.}
Recursive computation has usually served a different goal: deepening or
refining a single forward pass. The
Universal Transformer applies a shared block repeatedly across depth. Looped
Transformers study the resulting computation as a programmable process and as
iterative latent reasoning
\citep{dehghani2018universal,giannou2023looped,saunshi2502reasoning}.
Mixture-of-Recursions learns token-specific recursion depths inside a model
trained for recursive computation \citep{bae2026mixture}.

The same pattern appears in multimodal systems. ReVisE repeatedly updates
visual features, answers, and explanations \citep{ge2023wrong}. ReVisionLLM
narrows a temporal search interval over several passes for long-video grounding
\citep{hannan2025revisionllm}. ReT trains a recurrent cell for multimodal
document retrieval \citep{caffagni2025recurrence}. RecursiveVLM builds
recursion into continued pretraining and supervised fine-tuning through a
learned connector and a monotonic recursion loss \citep{xu2026looping}. In
these systems, repeated computation is learned for a particular refinement
task.

Closer to our setting, recent work rearranges or repeats the layers of
pretrained language models after training. CoLa searches, for each input, over
skipped, reordered, and repeated layers of a frozen LLM \citep{li2025skip}.
Inner Loop Inference repeatedly applies a selected block range
\citep{lys2026inner}. Training-Free Looped
Transformers instead uses damped substeps when looping a mid-stack block
\citep{chen2026trainingfree}. These studies show that selected blocks
of a frozen checkpoint can be reused at inference without retraining the base
model.

 Our work explores a complementary use of recursion as a source of candidate diversity. Each layer reuse configuration defines a distinct computation path for candidate generation, turning repeated computation into a sampling axis. Section~\ref{sec:architectural_sampling} formalizes the resulting candidate pool, which we evaluate for solution coverage and use as a source of rollouts for test-time learning. 

\input{method_final}

\input{results_final}
\input{ablation_final}
\input{conclusion}

\input{ai_statement}

\bibliography{iclr2027_conference}
\bibliographystyle{iclr2027_conference}

\appendix

\input{appendix}

\end{document}

%% file: math_commands.tex
\usepackage{amsmath,amsfonts,bm}

\def\eqref#1{equation~\ref{#1}}

\def\1{\bm{1}}

\DeclareMathAlphabet{\mathsfit}{\encodingdefault}{\sfdefault}{m}{sl}
\SetMathAlphabet{\mathsfit}{bold}{\encodingdefault}{\sfdefault}{bx}{n}



%% file: intro_try.tex
\definecolor{LavInk}{RGB}{92,64,158}   
\definecolor{BlueInk}{RGB}{36,81,166}  

\newcounter{resqcount}
\newcommand{\resq}[1]{%
  \stepcounter{resqcount}%
  \ifodd\value{resqcount}%
    \textcolor{LavInk}{\itshape #1}%
  \else
    \textcolor{BlueInk}{\itshape #1}%
  \fi
}

\section{Introduction}

Open-weight vision-language models (VLMs) are increasingly used exactly as they are shipped. A practitioner downloads a frozen checkpoint like Qwen-VL or Gemma and queries it directly, with no fine-tuning and no infrastructure beyond a forward pass \citep{bai2025qwen3, zhu2025internvl3}. In this regime the only way to extract more from a model is to spend more resources \emph{at inference}. This constraint has motivated a rapidly growing body
of work on test-time scaling, in which a model generates multiple candidate answers and an aggregation, verification, or selection procedure distills them into a stronger final solution~\citep{snell2025scaling, kaya2026efficient, singh2026ttrv}.

The promise of these methods rests on a single load-bearing assumption: that a VLM can produce a \emph{diverse} set of candidates that collectively provide a broader coverage of the solution space for downstream tasks. Yet this assumption conceals a fundamental question that the literature has largely left implicit: \resq{where does candidate diversity actually come from?}

The question matters because candidate-based test-time scaling is
fundamentally bounded by the candidates it is given. Generating $k$ answers is useful only when those answers provide meaningful coverage of the solution space. If all $k$ samples occupy essentially the same region of this space, increasing $k$ provides little additional value. The dominant way to create this diversity is remarkably simple: \emph{temperature sampling} \citep{holtzman2019curious}. However, temperature operates only at the final stage of the generation process. Every candidate is routed through the same layers with the same parameters and the same intermediate representations, and only the final draw from the resulting distribution differs. Diversity, in other words, is injected after all the interesting computation has already happened. Figure~\ref{fig:motivation} illustrates this contrast: standard-path temperature sampling varies the output draw after a common computation, whereas architectural sampling varies the computation that precedes it. 

This raises a natural follow-up: \resq{what if diversity need not come from changing the output distribution, but from changing the computation that produces it? Can a frozen model, without changing a single parameter, reveal new capabilities simply by exploring different ways of computing?}

We answer this question with \emph{architectural sampling}, illustrated in
Figure~\ref{fig:architectural_sampling_overview}. For each candidate, we select a short contiguous block of decoder layers, apply it a chosen number of times, and then resume the original forward pass. Changing the block location and repetition count yields several computation paths from one checkpoint. The weights remain frozen throughout, and the method adds no auxiliary parameters. Each path can be decoded greedily or with temperature sampling, allowing architectural and token-level variation to be combined.

This introduces a new axis for test-time scaling centered around architecture, in which candidate diversity arises from varying the computation performed by a frozen model. This raises a natural question: \resq{does architectural sampling reliably beat temperature sampling, and where do its gains come from?}

We observe consistent benefits across the VLM backbones and benchmarks
evaluated. Across five Qwen checkpoints and twelve multimodal
benchmarks, architectural sampling consistently improves $\mathrm{pass}@9$ over temperature sampling, with an average gain of approximately $6.58\%$. This advantage is not confined to a single budget: Figure~\ref{fig:qwen3_8b_scaling} traces $\mathrm{pass}@k$ for both methods as $k$ grows, showing that architectural sampling scales more favorably and dominates temperature sampling across the full range of candidate counts. The conditions behind these gains are consistent as well. Recursion is most effective when applied to early layers, while jointly updating vision and language tokens yields the strongest improvements. The gains even persist under greedy decoding, where repeated calls to the standard model return the same response. Architectural samples also have lower Self-BLEU than temperature samples, indicating lower lexical overlap \citep{self_bleu}.

Having established what the method does and where it is most effective, we are led to a final and more searching question: \resq{Beyond producing better samples, can architectural sampling actually improve how a model learns?}

To answer this, we use architectural sampling as a drop-in replacement for the temperature-based rollouts in TTRV, a label-free, GRPO-based test-time reinforcement learning method \citep{shao2024deepseekmath,singh2026ttrv}. Since TTRV learns directly from the model's own sampled responses, the quality and diversity of these candidates directly shape the resulting update. We find that replacing temperature-based rollouts with architectural rollouts consistently improves accuracy, showing that the benefits of architectural sampling extend beyond candidate generation and $\mathrm{pass}@k$ to test-time learning.

To conclude, we summarize the contributions of our work as follows:
\begin{itemize}
    \item We introduce architectural sampling, a training-free mechanism that reuses layer windows to produce candidates from a frozen VLM without modifying the checkpoint or adding parameters.
    \item We demonstrate the effectiveness of architectural sampling across five Qwen checkpoints and twelve multimodal benchmarks. At the same candidate budget, it improves $\mathrm{pass}@9$ by an average of 6.58\% points over temperature sampling.
   \item We further show that using architectural candidates as rollouts for label-free test-time reinforcement learning improves accuracy over temperature-sampled rollouts. These results extend the value of architectural sampling beyond candidate coverage, showing that it enables more effective learning from a model’s own outputs.
\end{itemize}

%% file: method_final.tex
\definecolor{LavInk}{RGB}{92,64,158}   
\definecolor{BlueInk}{RGB}{36,81,166}  

\newcommand{\blk}[1]{\textcolor{LavInk}{#1}}   
\newcommand{\rec}[1]{\textcolor{BlueInk}{#1}}  
\newcommand{\comp}{\circ}
\newcommand{\hh}[1]{h^{(#1)}}
\section{Architectural Sampling}
\label{sec:architectural_sampling}

\begin{figure}[t]
    \centering
    \includegraphics[width=1.0\linewidth]{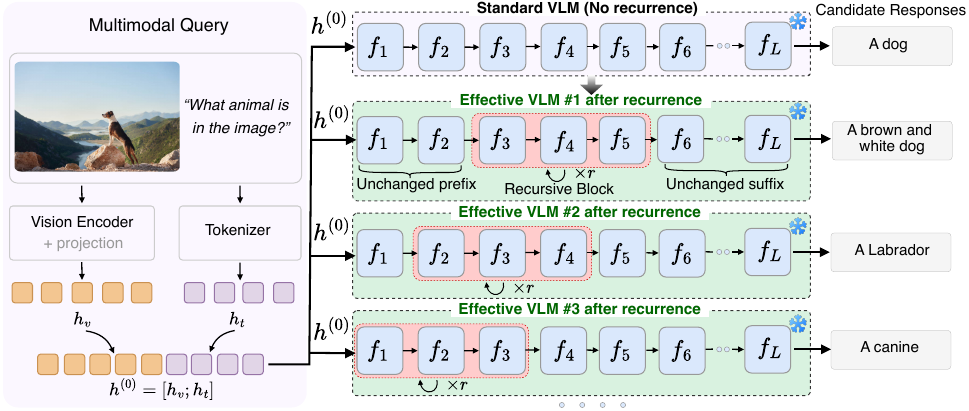}
    \caption{\captionheading{Architectural Sampling from a Frozen VLM}
    (left) The image and prompt form the initial multimodal context, to which
    generated tokens are appended. (right) The standard path applies every
    decoder layer once. Each architectural path repeats one contiguous window
    $\rec{r}$ times before continuing through the remaining frozen layers in their
    original order. All paths share frozen weights; changing the window and $\rec{r}$
    yields the candidate pool. Snowflakes mark frozen weights.}
    \label{fig:architectural_sampling_overview}
\end{figure}

Figure~\ref{fig:architectural_sampling_overview} shows how this sampling axis
changes the forward pass. Given the same image--prompt pair, every candidate
uses the same frozen checkpoint and decoding rule. The standard path applies
each decoder layer once. An architectural path instead revisits one contiguous
layer window before resuming the original layer order. Varying the window and
its recursion depth produces the candidate pool.

\subsection{Standard Forward Pass}

We consider a pretrained VLM that maps an input $x=(x_v,x_t)$ to a response
$y=(y_1,\ldots,y_m)$, where $x_v$ is an image and $x_t$ is a text prompt. The
vision encoder and projector map $x_v$ to visual tokens $h_v$, while the
tokenizer and embedding layer map $x_t$ to text embeddings $h_t$. At decoding
step $s$, the language decoder also receives embeddings $h_{y_{<s}}$ of the
tokens generated so far:
\begin{equation}
    h_s^{(0)} = [h_v;h_t;h_{y_{<s}}].
\end{equation}
For $s=1$, the generated prefix is empty. The decoder contains $L$ decoder
layers $f_1,\ldots,f_L$ with parameters $\theta_1,\ldots,\theta_L$. A standard
forward pass applies the layers once in their original order:
\begin{equation}
    h_s^{(\ell)} =
    f_\ell\!\left(h_s^{(\ell-1)};\theta_\ell\right),
    \qquad \ell=1,\ldots,L.
\end{equation}
The output head reads the final-position state in $h_s^{(L)}$ to predict $y_s$,
and this process repeats as the response grows. The checkpoint remains fixed
throughout candidate generation. Architectural sampling changes only this
computation path. To make that change explicit, we omit the step index $s$ and the fixed
parameters $\theta_\ell$ below, writing $\hh{0}=h_s^{(0)}$ for the decoder
input at the current step, $f_\ell(\cdot)$ for the frozen layer map, and
$\hh{\ell}$ for its output.

\subsection{Reusing a Layer Window}\label{method}

Architectural sampling perturbs the forward pass along two axes: \blk{which
window} of layers is chosen, and \rec{how many times} it is applied.
A \blk{block} is a
contiguous window of layers, and a \rec{recursion depth}~$\rec{r}$ re-applies
that window in place before the network proceeds:
\[
  \blk{B_{[a:b]}}\;=\;\blk{f_b\comp\cdots\comp f_a},
  \qquad\qquad
  \bigl(\blk{B_{[a:b]}}\bigr)^{\rec{\comp r}}
  \;=\;
  \underbrace{\blk{B_{[a:b]}}\comp\cdots\comp\blk{B_{[a:b]}}}_{\rec{r}\ \text{times}} .
\]
For a configuration $(\blk{B_{[a:b]}},\rec{r})$ the forward pass keeps the frozen
prefix and frozen suffix untouched and perturbs only the middle stage. Using the
same layer maps $f_\ell$ and hidden states $\hh{\ell}$ as the standard pass:
\begin{align*}
  \hh{a-1}
    &= f_{a-1}\comp\cdots\comp f_1\,\bigl(\hh{0}\bigr),
    && \text{\footnotesize frozen prefix (empty if }a=1)\\[3pt]
  \hh{b}
    &= \bigl(\blk{B_{[a:b]}}\bigr)^{\rec{\comp r}}\,\bigl(\hh{a-1}\bigr),
    && \text{\footnotesize reuse the }\blk{\text{window}}\ \rec{r}\ \text{\footnotesize times}\\[3pt]
  \hh{\ell}
    &= f_\ell\,\bigl(\hh{\ell-1}\bigr),\quad \ell=b{+}1,\dots,L,
    && \text{\footnotesize frozen suffix (empty if }b=L)
\end{align*}

Reading the whole pass left to right makes the logic explicit: only the
\blk{lavender} factor and its \rec{blue} exponent ever change, and the two
frozen stages stay fixed:
\[
  \hh{L}
  \;=\;
  \underbrace{f_L\comp\cdots\comp f_{b+1}}_{\text{\footnotesize suffix (fixed)}}
  \;\comp\;
  \underbrace{\bigl(\blk{B_{[a:b]}}\bigr)^{\rec{\comp r}}}_{\substack{\blk{\text{window}}\,\times\,\rec{r}\\[1pt]\text{\footnotesize (the two knobs)}}}
  \;\comp\;
  \underbrace{f_{a-1}\comp\cdots\comp f_1}_{\text{\footnotesize prefix (fixed)}}
  \bigl(\hh{0}\bigr).
\]

Setting $\rec{r}=1$ for any window recovers the standard forward pass exactly.
Increasing $\rec{r}$ re-processes the current representation through the same
frozen sub-network, while shifting $\blk{B_{[a:b]}}$ changes \emph{which}
computation is repeated. Recursion depth and window location thus induce
distinct feature trajectories through identical parameters, producing different
outputs from a single checkpoint.

\subsection{Building the Candidate Pool}

To build a candidate set of size $k$ we take the grid formed by a set of
\blk{windows}~$\blk{\mathcal{B}}$ and a set of \rec{recursion
depths}~$\rec{\mathcal{R}}$, and run one forward pass per pair, i.e.\ one
candidate per cell of the $\blk{\mathcal{B}}\!\times\!\rec{\mathcal{R}}$ grid:
\[
  \mathcal{Y}
  \;=\;
  \bigl\{\, y_{(\blk{B_{[a:b]}},\rec{r})} \;:\;
    \blk{B_{[a:b]}\in\mathcal{B}},\; \rec{r\in\mathcal{R}} \,\bigr\},
  \qquad
  k \;=\; \lvert\blk{\mathcal{B}}\rvert \cdot \lvert\rec{\mathcal{R}}\rvert .
\]

\noindent
These candidates support inference-time objectives such as $\mathrm{pass}@k$ and
any setting that requires a pool of self-generated trajectories, including RL
for VLMs. Because every configuration uses the same frozen checkpoint and the
same decoding temperature, gains reflect \emph{additional candidate diversity
induced by architectural variation}. Architectural sampling therefore provides a
general candidate-generation primitive, in which the choice of
$\blk{\mathcal{B}}$ and $\rec{\mathcal{R}}$ sets the size and diversity of the
resulting set $\mathcal{Y}$.

%% file: results_final.tex
\section{Results}
\label{sec:results}

The construction above turns one frozen checkpoint into a pool of nine
responses. We now ask whether this pool contains a correct answer more often
than an equal-sized pool drawn by temperature sampling along the standard path.

\subsection{Evaluation Settings}
\label{sec:evaluation_settings}

\paragraph{Models and benchmarks.}
We evaluate five Qwen checkpoints from three model generations: Qwen2.5-VL-7B
\citep{bai2025qwen25vl}, Qwen3-VL-2B and Qwen3-VL-8B
\citep{bai2025qwen3}, and Qwen3.5-VL-2B and Qwen3.5-VL-4B
\citep{qwen2026qwen35}. The main evaluation covers CV-Bench, RealWorldQA,
CountBench, MathVista, MMStar, and AI2D
\citep{tong2024cambrian,xai2024grok15v,paiss2023teaching,lu2023mathvista,chen2024mmstar,kembhavi2016diagram}.
We also include GQA, A-OKVQA, ScienceQA, CountQA, MMMU, and BLINK
\citep{hudson2019gqa,schwenk2022aokvqa,lu2022scienceqa,tamarapalli2025countqa,yue2023mmmu,fu2024blink}.
This set spans visual perception, counting, knowledge, and multimodal reasoning
in both multiple-choice and open-ended formats. Further details on the implementation and dataset characteristics are provided in Appendix~\ref{sec:appendix_implementation}.

\paragraph{Candidate generation.}
For architectural sampling, we use the pool defined in
Section~\ref{sec:architectural_sampling}. Following the one-based indexing used
there, the four layer windows are $[1{:}3]$, $[2{:}4]$, $[3{:}5]$, and
$[4{:}6]$. We generate one response from the standard path and one from each
window at recursion depths $r\in\{2,4\}$, giving nine candidates. Every path is
decoded with temperature $T=0.6$.

The baseline draws nine responses from the standard path, also with $T=0.6$.
Prompts, decoding settings, and answer evaluation are identical, so the
comparison matches candidate count and decoding temperature.

\paragraph{Evaluation Protocol.}
We report all results using the $\mathrm{pass}@k$ metric \citep{shao2024deepseekmath}. Given candidates
$\{y_1, \dots, y_k\}$, where $\mathbf{1}[\cdot]$ is the correctness indicator,
we define
\[
    \mathrm{pass}@k
    = \mathbb{E}\!\left[
        \max_{1 \le i \le k}
        \mathbf{1}[y_i \text{ is correct}]
      \right].
\]
\emph{i.e.,} a query is solved if at least one candidate is correct. Throughout the main Table~\ref{tab:main}, we evaluate with $k = 9$ and report $\mathrm{pass}@9$ as a percentage.

\subsection{Main Results}
\label{sec:main_results}

\input{tables/main_table}

Table~\ref{tab:main} compares architectural sampling with temperature sampling.
Across five checkpoints and twelve benchmarks, architectural sampling improves
$\mathrm{pass}@9$ by $6.58$ percentage points on average. The improvement is
consistent across all five checkpoint-level averages and occurs in $55$ of the
$60$ checkpoint--benchmark comparisons. The largest checkpoint-level gains are
observed for Qwen3-VL-8B ($+12.43$ points) and Qwen3.5-VL-4B ($+9.13$ points),
while the smallest gain is $+1.10$ points for Qwen3.5-VL-2B. The improvement
is also broad across tasks, with the largest
gains on MMMU ($+12.43$), MMStar ($+11.01$), BLINK ($+9.28$), and RealWorldQA
($+8.42$). Within each Qwen model family, the larger checkpoint consistently
benefits more, suggesting that increased model capacity leaves greater room for alternative computation paths to recover additional correct candidates. 

Importantly, this behavior is not specific to the Qwen family. We further
evaluate architectural sampling on InternVL3 and Gemma3 across a subset of
benchmarks, with results reported in Table~\ref{tab:generalization-family}.
Architectural sampling yields average gains of $1.54$ and $17.63$ percentage
points for InternVL3 and Gemma3, respectively. We additionally evaluate the
robustness of architectural sampling in Table~\ref{tab:variance}, where we
examine the consistency of its gains across repeated runs.
Taken together,
these results show that the benefits of varying the computation path extend
beyond a single model family and remain robust across different architectures,
model scales, and multimodal tasks.

%% file: tables/main_table.tex
\begin{table}[!t]
\centering
\setlength{\tabcolsep}{2pt}
\renewcommand{\arraystretch}{1.2}
\footnotesize
\caption{\captionheading{Candidate Coverage Across Checkpoints and Benchmarks}
$\mathrm{pass}@9$ for nine temperature samples from the standard path and nine
architectural candidates, one from each path, across five Qwen checkpoints and
twelve benchmarks. All candidates use $T=0.6$. The $\Delta$ row reports
architectural sampling minus standard-path temperature sampling. Avg. is the
mean across benchmarks, and the higher score for each
checkpoint--benchmark pair is bold.}
\label{tab:main}
\resizebox{0.99\linewidth}{!}{%
\begin{tabular}{@{}lccccccccccccc@{}}
\toprule
\rowcolor{HeaderNavy}
 &
\textbf{\textcolor{white}{CV-Bench}} & \textbf{\textcolor{white}{RealWorldQA}} &
\textbf{\textcolor{white}{CountBench}} & \textbf{\textcolor{white}{MathVista}} &
\textbf{\textcolor{white}{MMStar}} & \textbf{\textcolor{white}{AI2D}} &
\textbf{\textcolor{white}{GQA}} & \textbf{\textcolor{white}{A-OKVQA}} &
\textbf{\textcolor{white}{ScienceQA}} & \textbf{\textcolor{white}{CountQA}} &
\textbf{\textcolor{white}{MMMU}} & \textbf{\textcolor{white}{BLINK}} &
\cellcolor{AvgHead}\textbf{\textcolor{white}{Avg.}} \\
\midrule


\rowcolor{Group7B}
\multicolumn{13}{@{}l}{\textbf{\;Qwen2.5-VL-7B}} & \cellcolor{AvgCell} \\
\rowcolor{Band7B}
Temperature & 92.46 & 86.67 & 97.75 & 72.60 & 79.47 & 90.75 & 70.15 & 93.36 & 92.71 & 86.36 & 78.28 & 87.52 & \avg{85.67} \\
\rowcolor{Band7Bours}
Architectural & \best{95.72} & \best{89.80} & \best{98.82} & \best{75.20} & \best{86.80} & \best{95.52} & \best{76.50} & \best{97.03} & \best{97.37} & \best{86.99} & \best{83.15} & \best{94.07} & \avg{\best{89.75}} \\
\rowcolor{Band7B}
$\Delta$ & \gain{3.26} & \gain{3.13} & \gain{1.07} & \gain{2.60} & \gain{7.33} & \gain{4.77} & \gain{6.35} & \gain{3.67} & \gain{4.66} & \gain{0.63} & \gain{4.87} & \gain{6.55} & \avg{\gain{4.07}} \\
\midrule

\rowcolor{Group3B}
\multicolumn{13}{@{}l}{\textbf{\;Qwen3-VL-2B}} & \cellcolor{AvgCell} \\
\rowcolor{Band3B}
Temperature & 85.22 & 72.29 & 94.62 & 43.70 & 53.33 & 74.85 & 64.60 & 85.68 & 75.66 & 82.30 & 35.33 & 63.43 & \avg{69.25} \\
\rowcolor{Band3Bours}
Architectural & \best{89.58} & \best{85.23} & \best{96.94} & \best{50.90} & \best{60.20} & \best{76.83} & \best{70.70} & \best{88.91} & \best{81.01} & \best{85.84} & \best{48.81} & \best{69.86} & \avg{\best{75.40}} \\
\rowcolor{Band3B}
$\Delta$ & \gain{4.36} & \gain{12.94} & \gain{2.32} & \gain{7.20} & \gain{6.87} & \gain{1.98} & \gain{6.10} & \gain{3.23} & \gain{5.35} & \gain{3.54} & \gain{13.48} & \gain{6.43} & \avg{\gain{6.15}} \\
\midrule

\rowcolor{Group7B}
\multicolumn{13}{@{}l}{\textbf{\;Qwen3-VL-8B}} & \cellcolor{AvgCell} \\
\rowcolor{Band7B}
Temperature & 88.48 & 72.94 & 97.27 & 58.70 & 58.93 & 84.89 & 64.75 & 88.65 & 88.75 & 76.52 & 48.56 & 71.50 & \avg{75.00} \\
\rowcolor{Band7Bours}
Architectural & \best{96.40} & \best{88.10} & \best{98.83} & \best{64.10} & \best{82.73} & \best{93.48} & \best{76.20} & \best{95.72} & \best{97.97} & \best{86.05} & \best{80.27} & \best{89.28} & \avg{\best{87.43}} \\
\rowcolor{Band7B}
$\Delta$ & \gain{7.92} & \gain{15.16} & \gain{1.56} & \gain{5.40} & \gain{23.80} & \gain{8.59} & \gain{11.45} & \gain{7.07} & \gain{9.22} & \gain{9.53} & \gain{31.71} & \gain{17.78} & \avg{\gain{12.43}} \\
\midrule

\rowcolor{Group3B}
\multicolumn{13}{@{}l}{\textbf{\;Qwen3.5-VL-2B}} & \cellcolor{AvgCell} \\
\rowcolor{Band3B}
Temperature & 91.70 & 85.75 & 98.08 & \best{65.60} & 73.47 & \best{92.03} & 71.25 & 91.79 & \best{92.02} & 82.14 & \best{59.93} & 80.20 & \avg{82.00} \\
\rowcolor{Band3Bours}
Architectural & \best{94.20} & \best{92.94} & \best{98.46} & 63.70 & \best{74.27} & 85.58 & \best{78.40} & \best{94.41} & 90.68 & \best{87.16} & 55.93 & \best{81.46} & \avg{\best{83.10}} \\
\rowcolor{Band3B}
$\Delta$ & \gain{2.50} & \gain{7.19} & \gain{0.38} & \drop{1.90} & \gain{0.80} & \drop{6.45} & \gain{7.15} & \gain{2.62} & \drop{1.34} & \gain{5.02} & \drop{4.00} & \gain{1.26} & \avg{\gain{1.10}} \\
\midrule

\rowcolor{Group7B}
\multicolumn{13}{@{}l}{\textbf{\;Qwen3.5-VL-4B}} & \cellcolor{AvgCell} \\
\rowcolor{Band7B}
Temperature & 68.42 & 88.50 & 97.47 & 64.80 & 42.60 & 49.18 & 72.20 & 87.34 & 63.71 & \best{77.34} & 30.21 & 63.05 & \avg{67.07} \\
\rowcolor{Band7Bours}
Architectural & \best{91.39} & \best{92.16} & \best{98.09} & \best{66.80} & \best{58.87} & \best{65.24} & \best{76.85} & \best{88.82} & \best{75.71} & 76.73 & \best{46.32} & \best{77.43} & \avg{\best{76.20}} \\
\rowcolor{Band7B}
$\Delta$ & \gain{22.97} & \gain{3.66} & \gain{0.62} & \gain{2.00} & \gain{16.27} & \gain{16.06} & \gain{4.65} & \gain{1.48} & \gain{12.00} & \drop{0.61} & \gain{16.11} & \gain{14.38} & \avg{\gain{9.13}} \\

\bottomrule
\end{tabular}%
}
\end{table}

%% file: ablation_final.tex
\section{Understanding Architectural Sampling}
\label{sec:analysis}

Having established that architectural sampling improves inference-time performance, we now examine it more closely. Unless noted otherwise, all experiments in this section use Qwen3-VL-8B on a five-dataset subset of our benchmarks. With the model and data held fixed, we study the roles of layers and tokens, connect our results to broader questions of temperature and diversity, and conclude with a practical demonstration that further strengthens our claims about its success.

\paragraph{Where should the window be placed?}
\input{tables/layers_token_table_w_o_del}
Table~\ref{tab:ablation} moves the repeated window from early to middle and late decoder layers while keeping the candidate budget fixed.
Every region improves over temperature sampling, but the gain falls as the
window moves deeper. Reusing the early layers gives the highest
$\mathrm{pass}@9$ on all five benchmarks, so we use early windows in the main configuration. 

\paragraph{Which tokens should be reprocessed?}
Table~\ref{tab:ablation-tokens} compares different token update strategies within the repeated window. Jointly reprocessing visual and language tokens achieves the highest $\mathrm{pass}@9$ on all five benchmarks. Language-only updates retain most of the improvement, while updating only visual tokens or their averaged representation yields smaller gains. We therefore keep the reprocessing of both modalities in our main configuration.  

\begin{figure}[!t]
    \centering
    \begin{subfigure}[t]{0.39\linewidth}
        \centering
        \includegraphics[width=\linewidth]{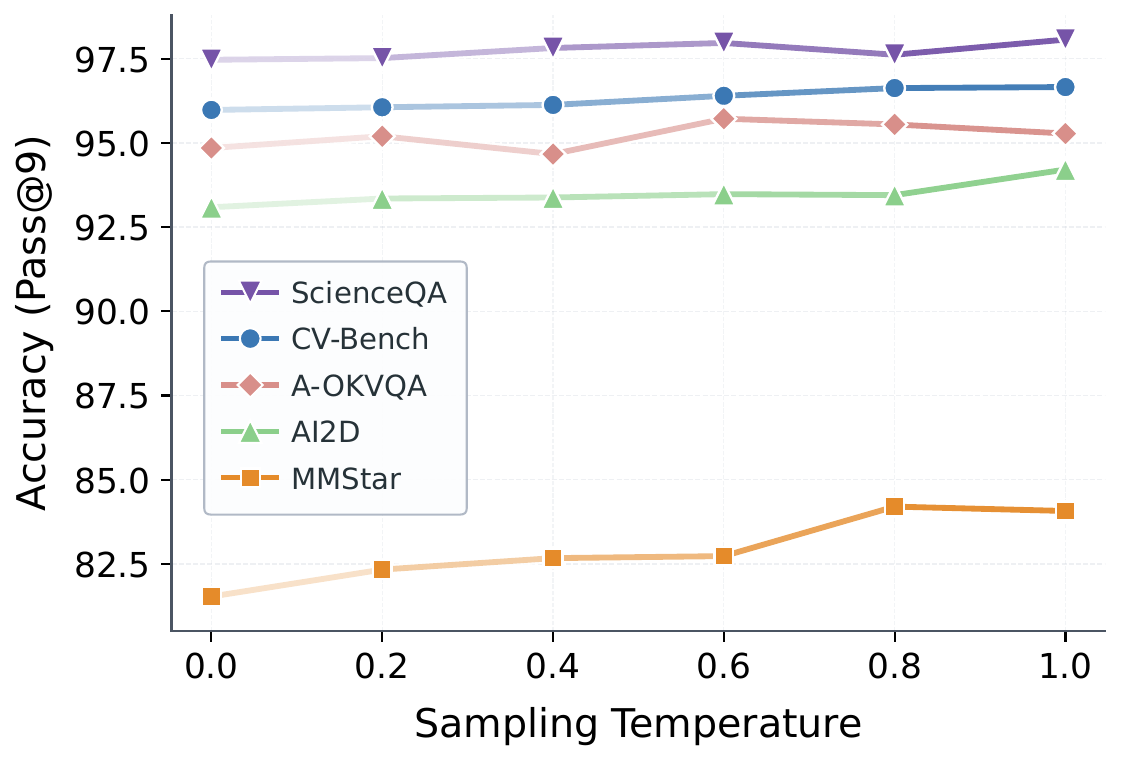}
        \caption{Effect of decoding temperature on architectural sampling.}
        \label{fig:temperature_sweep}
    \end{subfigure}\hfill
    \begin{subfigure}[t]{0.59\linewidth}
        \centering
        \includegraphics[width=\linewidth]{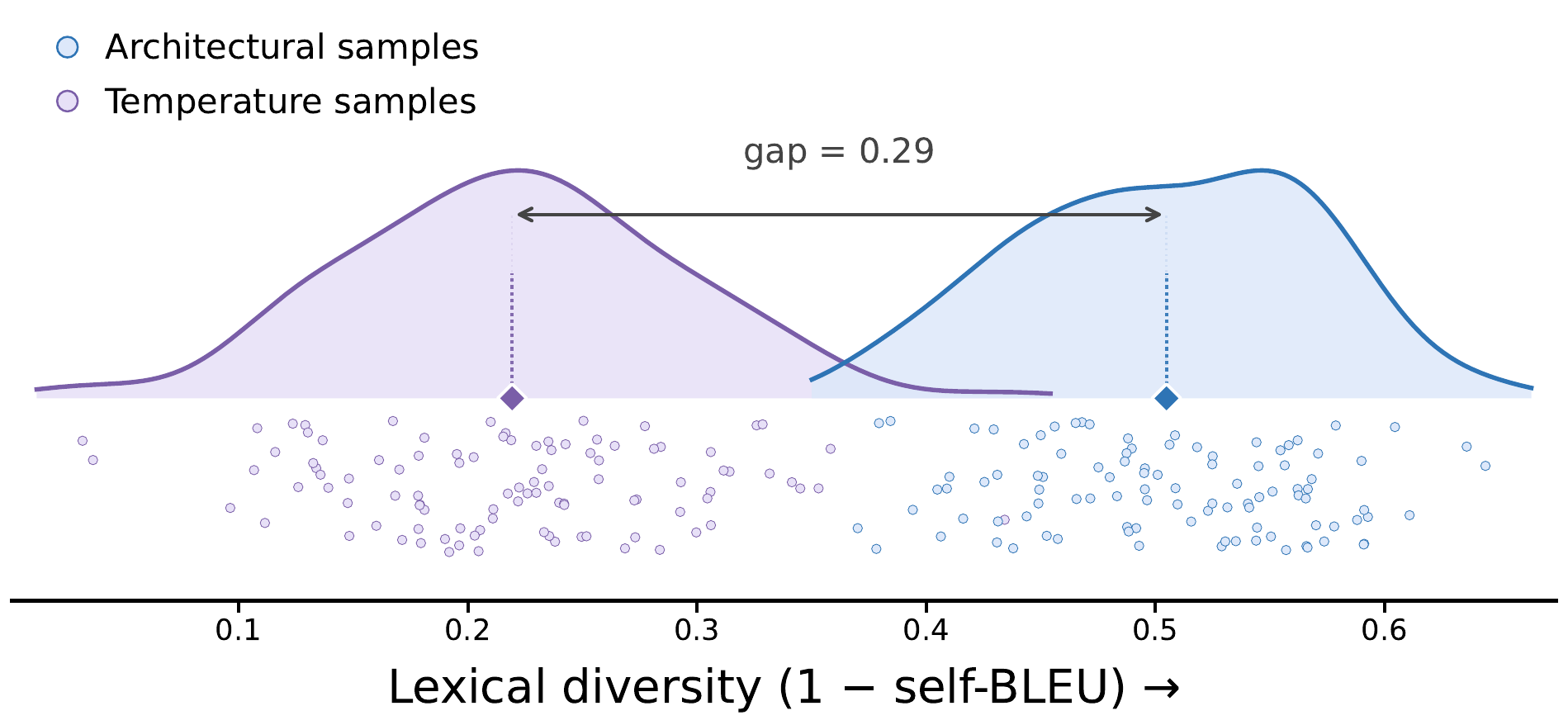}
        \caption{Lexical diversity of the two candidate pools.}
        \label{fig:lexical_diversity}
    \end{subfigure}
    \caption{\captionheading{Temperature Robustness and Candidate Diversity}
    Results on Qwen3-VL-8B over the five benchmarks. (left)
     Architectural sampling maintains stable $\mathrm{pass}@9$ across decoding temperatures, including greedy decoding ($T=0$). (right) Architectural candidates have higher
    $1-\mathrm{Self\text{-}BLEU}$ than temperature samples. Density
curves and individual observations show the two distributions, with diamonds marking their means ($0.51$ vs.\ $0.22$).}
    \label{fig:temperature_and_diversity}
\end{figure}
\input{tables/decoding_tec}

\paragraph{How does architectural sampling interact with temperature?}
Architectural sampling can be applied at any decoding temperature, including temperature $0$, where decoding is fully greedy. As shown in Figure~\ref{fig:temperature_sweep}, accuracy remains largely stable across the full range of temperatures we evaluate. This suggests that architectural sampling provides a consistent source of candidate diversity that is largely independent of decoding temperature, maintaining broad coverage of the answer space even at low temperatures.

\paragraph{How diverse are the responses?}
We measure lexical diversity using $1-\mathrm{Self\text{-}BLEU}$ \citep{self_bleu}, where higher values indicate lower lexical overlap among candidates. Figure~\ref{fig:lexical_diversity} shows a substantially higher mean diversity for architectural sampling ($0.51$) than for temperature sampling ($0.22$). Architectural candidates therefore exhibit considerably less lexical overlap and explore a broader range of responses.

\paragraph{Does architectural sampling outperform standard decoding strategies?}

Table~\ref{tab:decoding} compares architectural sampling with beam search, top-$k$, top-$p$, and temperature sampling at the same candidate budget. Architectural sampling achieves the highest $\mathrm{pass}@9$ across all benchmarks in this comparison, indicating stronger candidate coverage than standard decoding strategies. These results show that varying the final output distribution alone provides only limited gains, since all standard decoding strategies sample from outputs produced by the same underlying computation. In contrast, architectural sampling varies the computation itself, leading to substantially greater candidate diversity and, consequently, higher $\mathrm{pass}@9$. This highlights the value of introducing diversity upstream in the computation rather than only at the final decoding stage.

\paragraph{Can architectural candidates improve test-time learning?}
A diverse candidate pool is valuable not only for inference but also for self-improvement methods that learn from a model's own outputs. To test this, we adopt a label-free reinforcement-learning setup in the style of TTRV \citep{singh2026ttrv}, where the reward is derived from the model's own architectural rollouts rather than ground-truth labels, and we update the model with GRPO \citep{shao2024deepseekmath}. We evaluate on two backbones, Qwen3-VL-8B and Qwen3.5-VL-2B, across six datasets, comparing standard TTRV (temperature-sampled rollouts) against our variant using architectural rollouts, measured by standard accuracy or $\mathrm{pass}@1$ as shown in Table~\ref{tab:ttrv}. Architectural sampling consistently improves over vanilla TTRV across backbones and datasets, showing that its induced diversity is not mere noise: it carries a signal correlated with correctness that gradient-based optimization can exploit. This demonstrates that architectural sampling is useful not only for test-time inference but also as a source of informative self-supervision for test-time training.

\input{tables/ttrv}
\FloatBarrier


%% file: tables/layers_token_table_w_o_del.tex
\begin{table}[!t]
\centering
\setlength{\tabcolsep}{6pt}
\renewcommand{\arraystretch}{1.3}
\footnotesize

\begin{minipage}[t]{0.49\textwidth}
\centering
\caption{\captionheading{Layer-Window Placement} $\mathrm{pass}@9$ when the
repeated window is placed in early, middle, or late layers of Qwen3-VL-8B. The
best setting per benchmark is bold.}
\label{tab:ablation}
\resizebox{0.96\linewidth}{!}{%
\begin{tabular}{@{}lccccc@{}}
\toprule
\rowcolor{HeaderNavy}
 &
\textbf{\textcolor{white}{CV-Bench}} & \textbf{\textcolor{white}{MMStar}} &
\textbf{\textcolor{white}{AI2D}} & \textbf{\textcolor{white}{A-OKVQA}} &
\textbf{\textcolor{white}{ScienceQA}} \\
\midrule
\rowcolor{AvgCell}
Temperature & 88.48 & 58.93 & 84.89 & 88.65 & 88.75 \\
\midrule
\rowcolor{Band7Bours}
Early layers &
\best{96.40} & \best{82.73} & \best{93.48} & \best{95.72} & \best{97.97} \\
Middle layers &
93.25 & 76.13 & 92.96 & 92.05 & 95.04 \\
Late layers &
90.83 & 67.20 & 88.28 & 90.57 & 92.96 \\
\bottomrule
\end{tabular}%
}
\end{minipage}
\hfill
\begin{minipage}[t]{0.49\textwidth}
\centering
\caption{\captionheading{Tokens Reprocessed by the Window}
$\mathrm{pass}@9$ when the repeated window updates different token states in
Qwen3-VL-8B. The best setting per benchmark is bold.}
\label{tab:ablation-tokens}
\resizebox{0.96\linewidth}{!}{%
\begin{tabular}{@{}lccccc@{}}
\toprule
\rowcolor{HeaderNavy}
 &
\textbf{\textcolor{white}{CV-Bench}} & \textbf{\textcolor{white}{MMStar}} &
\textbf{\textcolor{white}{AI2D}} & \textbf{\textcolor{white}{A-OKVQA}} &
\textbf{\textcolor{white}{ScienceQA}} \\
\midrule
\rowcolor{AvgCell}
Temperature & 88.48 & 58.93 & 84.89 & 88.65 & 88.75 \\
\midrule
Only visual tokens &
90.60 & 65.73 & 90.85 & 91.80 & 94.10 \\
Only language tokens &
96.29 & 81.60 & 93.42 & 95.11 & 96.68 \\
Average visual token &
89.61 & 62.20 & 88.45 & 90.04 & 92.61 \\
\rowcolor{Band7Bours}
All tokens &
\best{96.40} & \best{82.73} & \best{93.48} & \best{95.72} & \best{97.97} \\
\bottomrule
\end{tabular}%
}
\end{minipage}

\end{table}

%% file: tables/decoding_tec.tex
\begin{table}[!t]
\centering
\setlength{\tabcolsep}{6pt}
\renewcommand{\arraystretch}{1.3}
\footnotesize
\caption{\captionheading{Comparison with Standard Decoding strategies}
$\mathrm{pass}@9$ for Qwen3-VL-8B using temperature sampling, beam search,
top-$k$, top-$p$, and architectural sampling under equal candidate budgets.
Architectural sampling is highlighted; the best score per benchmark is bold.}
\label{tab:decoding}
\resizebox{\ifdim\width>\linewidth \linewidth\else \width\fi}{!}{%
\begin{tabular}{@{}lccccc@{}}
\toprule
\rowcolor{HeaderNavy}
 &
\textbf{\textcolor{white}{CV-Bench}} & \textbf{\textcolor{white}{MMStar}} &
\textbf{\textcolor{white}{AI2D}} & \textbf{\textcolor{white}{A-OKVQA}} &
\textbf{\textcolor{white}{ScienceQA}} \\
\midrule
Temperature & 88.48 & 58.93 & 84.89 & 88.65 & 88.75 \\
Beam Search & 86.09 & 52.60 & 80.78 & 87.16 & 86.22 \\
Top-$k$  & 89.12 & 60.27 & 85.65 & 89.61 & 89.74 \\
Top-$p$ & 88.91 & 59.84 & 85.37 & 89.28 & 89.41 \\
\midrule
\rowcolor{Band7Bours}
Architectural &
\best{96.40} & \best{82.73} & \best{93.48} & \best{95.72} & \best{97.97} \\
\bottomrule
\end{tabular}%
}
\end{table}

%% file: tables/ttrv.tex
\begin{table}[H]
\centering
\setlength{\tabcolsep}{6pt}
\renewcommand{\arraystretch}{1.1}
\footnotesize
\caption{\captionheading{Architectural Rollouts for Test-Time Learning}
$\mathrm{pass}@1$ accuracy after TTRV on Qwen3.5-VL-2B and Qwen3-VL-8B. Architectural-rollout cells report the absolute change over temperature-sampled rollouts; the best setting per benchmark is shown in bold.}
\label{tab:ttrv}
\resizebox{0.96\linewidth}{!}{%
\begin{tabular}{@{}lcccccc@{}}
\toprule
\rowcolor{HeaderNavy}
 &
\textbf{\textcolor{white}{CV-Bench}} & \textbf{\textcolor{white}{MMStar}} &
\textbf{\textcolor{white}{AI2D}} & \textbf{\textcolor{white}{A-OKVQA}} &
\textbf{\textcolor{white}{ScienceQA}} & \textbf{\textcolor{white}{CountBench}} \\
\midrule
\rowcolor{Group3B}
\multicolumn{7}{@{}l@{}}{\textbf{\textsc{Qwen3.5-VL-2B}}} \\
\midrule
\rowcolor{Band3B}
Temperature rollouts &
74.60 & 45.52 & 75.38 & 68.04 & 67.01 & 93.06 \\
\rowcolor{Band3Bours}
Architectural rollouts &
\best{80.29}\,\gain{5.69} & \best{49.17}\,\gain{3.66} & \best{76.94}\,\gain{1.56} &
\best{83.38}\,\gain{15.34} & \best{77.33}\,\gain{10.32} & \best{95.10}\,\gain{2.04} \\
\midrule
\rowcolor{Group7B}
\multicolumn{7}{@{}l@{}}{\textbf{\textsc{Qwen3-VL-8B}}} \\
\midrule
\rowcolor{Band7B}
Temperature rollouts &
86.17 & 53.93 & 76.61 & 87.40 & 84.75 & 93.17 \\
\rowcolor{Band7Bours}
Architectural rollouts &
\best{86.29}\,\gain{0.11} & \best{55.52}\,\gain{1.59} & \best{77.80}\,\gain{1.17} &
\best{87.95}\,\gain{0.55} & \best{88.82}\,\gain{4.07} & \best{96.93}\,\gain{3.76} \\
\bottomrule
\end{tabular}%
}
\end{table}

%% file: conclusion.tex
\section{Limitations and Conclusion}
\label{sec:conclusion}

\paragraph{Limitations.}
Our method comes with two limitations. First, because architectural sampling relies on recursion at its core, it increases the per-candidate forward cost relative to temperature sampling, where every candidate consumes the same resources. We provide an extensive cost analysis in Appendix~\ref{sec:cost} and find that, with the careful use of recursion as adopted in our work, the overhead is marginal (roughly 1--2 seconds more, depending on recursion depth, up to depth 4). Nevertheless, this added cost may matter in resource-constrained settings, and we encourage readers to weigh it accordingly. Second, architectural sampling requires access to the internal decoder layers, which precludes its use with closed-weight models served only through an API.

\paragraph{Conclusion.}
We introduced architectural sampling, a training-free way to generate
candidates from different computation paths through a frozen VLM. Each path
reuses an early layer window before resuming the original forward pass, turning
one checkpoint into a family of candidate generators. At a fixed budget of nine
candidates, this improves $\mathrm{pass}@9$ over temperature sampling by 6.58
percentage points on average across our main evaluation. The ablations show
that early multimodal reprocessing works best among the tested settings, while
coverage remains high under greedy decoding and the resulting candidates have
less lexical overlap. When used as rollouts, these candidates also improve
test-time learning. Architectural sampling therefore closes the loop on our opening questions: a frozen VLM need not follow a single computation path at
test time. Its own computation graph provides a second sampling space,
complementary to token-level randomness and available without retraining the
checkpoint. This reframes what test-time scaling can mean. We have been scaling test-time compute by asking a model to say more things, when we should have been letting it think in more ways. The weights, it turns out, were never the ceiling. The ceiling was our insistence on using them in only one way.

%% file: ai_statement.tex

\subsection*{AI Use Statement}


Generative AI tools assisted with language editing, manuscript organization,
and LaTeX presentation. They were not used to generate experimental data or
determine the reported results. The authors reviewed the resulting text,
equations, citations, and presentation, and take responsibility for the final
manuscript.

\subsection*{Reproducibility Statement}

The main paper and appendix summarize the architectural configurations,
candidate budget, decoding settings, model families, benchmarks, and metrics
used in our experiments. We will release the implementation and evaluation
scripts with the final paper.

%% file: appendix.tex
\newpage
\section{Appendix}

In this appendix, we present additional experiments and implementation details that provide further
insight into our method and complement the results presented in the main manuscript. Section~\ref{sec:appendix_implementation}
provides additional implementation details for our experiments. Section~\ref{sec:appendix_ablation} presents additional ablation studies that examine further aspects of our method and provide deeper insights into its effectiveness. Section~\ref{sec:appendix_scaling} presents additional scaling curves corresponding to the results reported
in the main tables. Finally, Section~\ref{sec:cost} provides the computational cost analysis of our method.
All experiments were conducted on a machine equipped with an NVIDIA A100 GPU. To facilitate
reproducibility during the review process, we also provide the complete codebase (code.zip) together with detailed execution instructions in README.md. The complete codebase will be publicly
released upon acceptance.

\section{Additional Implementation Details}
\label{sec:appendix_implementation}

\paragraph{Inference-time sampling.}
All checkpoint parameters remain frozen during candidate generation. We use
the four inclusive windows $[1{:}3]$, $[2{:}4]$, $[3{:}5]$, and $[4{:}6]$,
together with recursion depths $r\in\{2,4\}$ and one standard path. This gives
nine candidates per input. Each path is decoded with temperature $T=0.6$. We
generate at most 32 new tokens for multiple-choice questions and 256 for
open-ended questions. The temperature baseline uses the same limits and draws
nine responses from the standard path.

\paragraph{Test-time learning.}
For the experiment in Table~\ref{tab:ttrv}, we retain the label-free,
GRPO-based TTRV procedure \citep{shao2024deepseekmath,singh2026ttrv} and replace
its temperature-sampled rollouts with the architectural candidates. For
each training prompt we form the rollout group of $6$ from these variants, each
decoded at temperature $0.6$, and assign the scalar reward by majority
vote. No ground-truth labels are therefore used during
training. Group-relative advantages are computed with the standard
mean/standard-deviation normalization. We train on only $20$ samples and evaluate on all remaining held-out data. The policy is adapted with LoRA (rank $r{=}32$,
$\alpha{=}64$, dropout $0.05$) applied to the attention and MLP projection layers
of the language backbone, with the vision encoder kept frozen. Optimization uses AdamW at a learning rate of $1{\times}10^{-5}$ in \texttt{bf16}, with gradient clipping at $1.0$ and a KL penalty ($\beta{=}0.02$) toward a frozen reference policy obtained by disabling the LoRA adapter. We report greedy pass@1.

\paragraph{Dataset Description.} To comprehensively evaluate our approach, we curate a diverse set of multimodal benchmarks spanning different aspects of visual understanding and reasoning. The complete list of datasets considered in our evaluation, along with their corresponding characteristics, is provided in Table~\ref{tab:dataset}. Our selection is designed to cover both fundamental visual capabilities and higher-level reasoning, allowing us to assess the generalization of our approach across heterogeneous tasks and evaluation settings.

\begin{table}[!ht]
\centering
\renewcommand{\arraystretch}{1.0}
\begin{adjustbox}{max width=0.97\textwidth}
\begin{tabular}{@{}
    >{\raggedright\arraybackslash}p{2.7cm}
    >{\raggedright\arraybackslash}p{6.4cm}
    >{\raggedright\arraybackslash}p{4.6cm}@{}}
\toprule
\textbf{Dataset} & \textbf{Dataset Focus} & \textbf{Dataset Type} \\
\midrule

\rowcolor{lightgray}
AI2D~\citep{kembhavi2016diagram}
    & Diagram understanding and reasoning
    & MCQ \\

ScienceQA~\citep{lu2022scienceqa}
    & Multimodal science QA and reasoning
    & MCQ \\

\rowcolor{lightgray}
CV-Bench~\citep{tong2024cambrian}
    & 2D/3D visual understanding (spatial relations, counting, depth \& distance)
    & MCQ \\

MMStar~\citep{chen2024mmstar}
    & Multimodal perception and reasoning (vision-indispensable)
    & MCQ \\

\rowcolor{lightgray}
BLINK~\citep{fu2024blink}
    & Core visual perception abilities (relative depth, correspondence, multi-view, etc.)
    & MCQ (with visual prompts) \\

\midrule

CountBench~\citep{paiss2023teaching}
    & Object counting
    & Numeric \\

\rowcolor{lightgray}
CountQA~\citep{tamarapalli2025countqa}
    & Object counting in real-world scenes (high density, clutter, occlusion)
    & Numeric / short-answer \\

\midrule

GQA~\citep{hudson2019gqa}
    & Real-world visual reasoning and compositional QA
    & Mixed: binary + open-ended \\

\rowcolor{lightgray}
A-OKVQA~\citep{schwenk2022aokvqa}
    & Visual QA with world knowledge
    & Mixed: MCQ + direct answer \\

MathVista~\citep{lu2023mathvista}
    & Mathematical reasoning in visual contexts
    & Mixed: MCQ + free-form \\

\rowcolor{lightgray}
MMMU~\citep{yue2023mmmu}
    & Multidisciplinary multimodal understanding and reasoning
    & Mixed: MCQ + open-ended \\

RealWorldQA~\citep{xai2024grok15v}
    & Real-world spatial understanding
    & Mixed: MCQ + numeric + open-ended \\

\bottomrule
\end{tabular}
\end{adjustbox}
\caption{\textbf{Statistics of the visual reasoning and VQA datasets used in our experiments.}
We report the primary focus and dataset type for each benchmark, covering a diverse range of visual perception, reasoning, counting, mathematical, and real-world understanding tasks.}
\label{tab:dataset}
\end{table}

We first consider benchmarks that primarily evaluate \textit{visual perception} and \textit{structured reasoning}. These include AI2D~\citep{kembhavi2016diagram}, ScienceQA~\citep{lu2022scienceqa}, CV-Bench~\citep{tong2024cambrian}, MMStar~\citep{chen2024mmstar}, and BLINK~\citep{fu2024blink}. Together, these benchmarks provide complementary evaluations of models' ability to interpret visual content, understand spatial relationships, reason over structured visual information, and perform vision-dependent tasks.

To examine more specific reasoning capabilities, we include CountBench~\citep{paiss2023teaching} and CountQA~\citep{tamarapalli2025countqa}, which evaluate \textit{visual counting} under increasingly challenging scene conditions. We further include GQA~\citep{hudson2019gqa}, A-OKVQA~\citep{schwenk2022aokvqa}, and MMMU~\citep{yue2023mmmu} to assess broader \textit{multimodal reasoning}, including compositional understanding and reasoning that may require information beyond the immediate visual content. Finally, MathVista~\citep{lu2023mathvista} and RealWorldQA~\citep{xai2024grok15v} extend the evaluation toward \textit{mathematical} and \textit{real-world reasoning}. These benchmarks require models to combine visual information with reasoning processes and, in some cases, external knowledge or domain-specific understanding.

Overall, the benchmark suite in Table~\ref{tab:dataset} provides a broad and complementary evaluation of multimodal capabilities. By combining general visual understanding benchmarks with targeted evaluations of \textit{counting}, \textit{spatial reasoning}, \textit{mathematical reasoning}, and \textit{higher-level multimodal understanding}, we evaluate our approach across a wide range of challenging visual reasoning scenarios.

\section{Additional Experiments and Ablations}
\label{sec:appendix_ablation}

\paragraph{Robustness.} To measure run-to-run variation, Table~\ref{tab:variance} repeats the Qwen3-VL-8B comparison five times. The standard deviation remains below $0.32$ points for both candidate-generation strategies on all five benchmarks.
\input{tables/variance}

\paragraph{Generalization to model families.} The main evaluation uses Qwen checkpoints \citep{bai2025qwen25vl,bai2025qwen3,qwen2026qwen35}. Table~\ref{tab:generalization-family} adds InternVL3-2B \citep{zhu2025internvl3} and Gemma3-4B \citep{team2025gemma} on the same five-benchmark subset. Architectural sampling raises the five-benchmark average by $1.54$ points for
InternVL3-2B and $17.63$ points for Gemma3-4B. These results provide an initial
test beyond Qwen. Broader model-family coverage remains to be studied.

\input{tables/model_families}
\FloatBarrier

\section{Scaling Curves}
\label{sec:appendix_scaling}

\begin{figure}[H]
\centering
\includegraphics[width=0.8\linewidth]{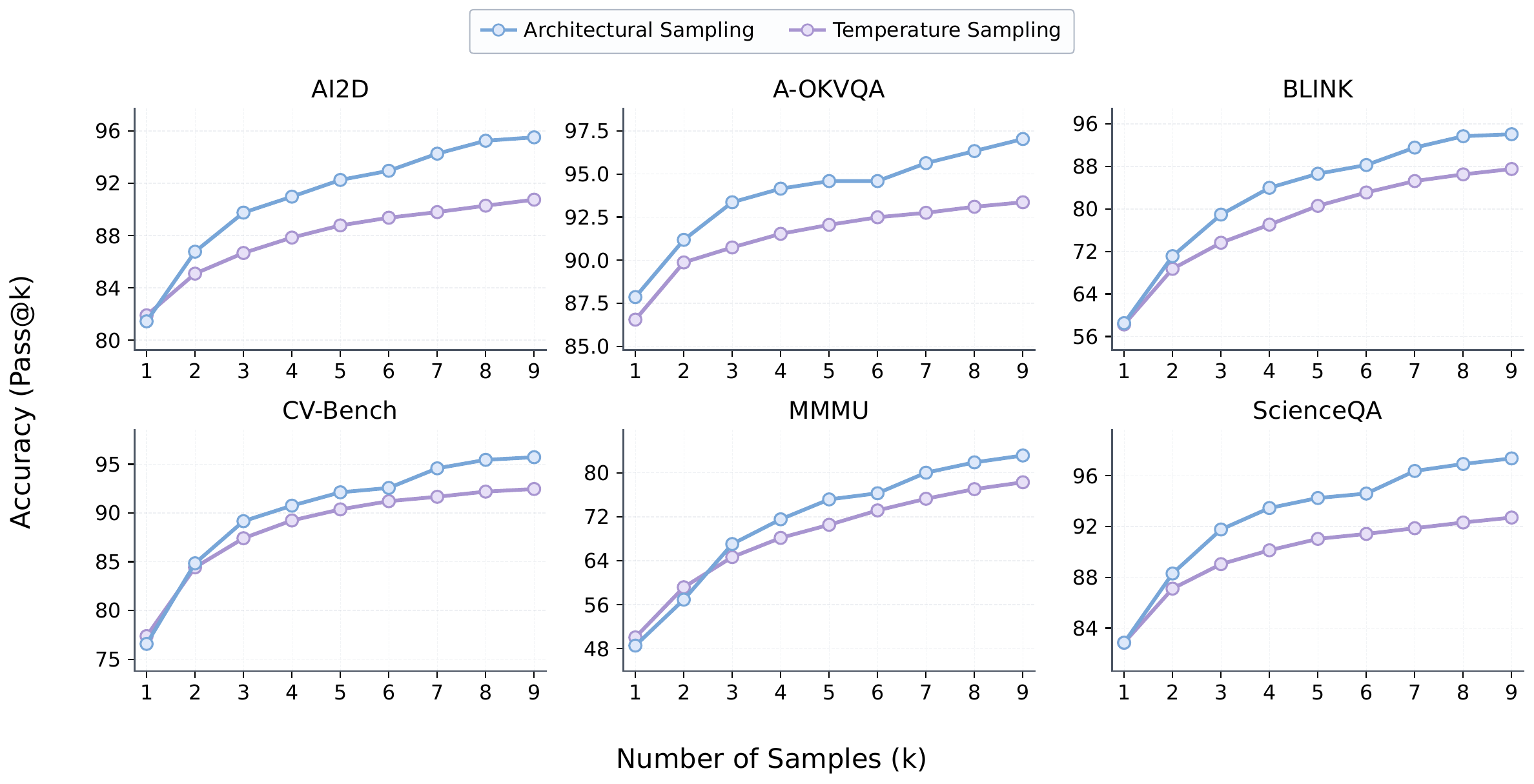}
\caption{\captionheading{Candidate Coverage as the Pool Grows: Qwen2.5-VL-7B}
$\mathrm{pass}@k$ across six benchmarks as the candidate budget grows for
architectural and temperature sampling.}
\label{fig:qwen25_7b_scaling}
\end{figure}

\begin{figure}[H]
\centering
\includegraphics[width=0.8\linewidth]{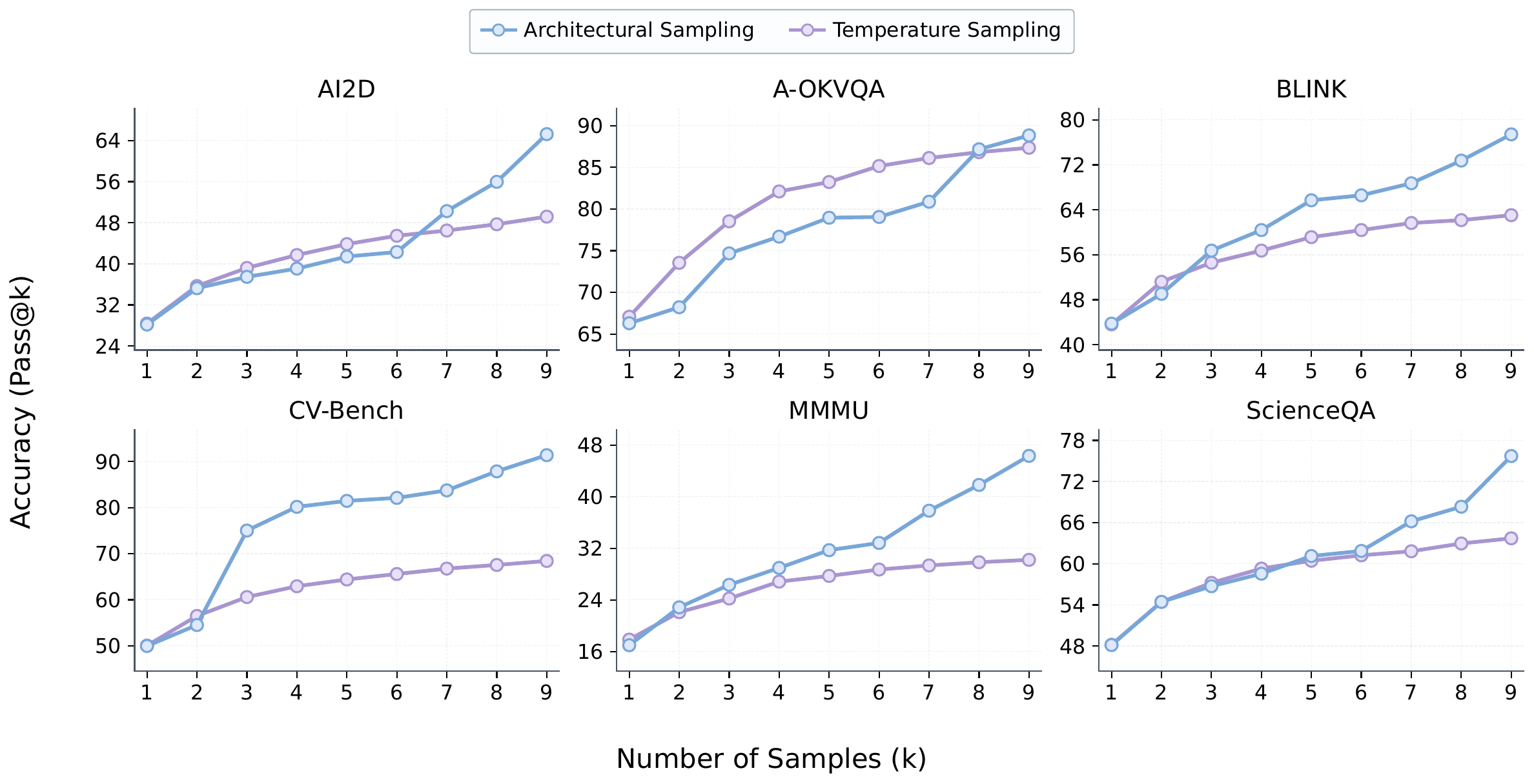}
\caption{\captionheading{Candidate Coverage as the Pool Grows: Qwen3.5-VL-4B}
$\mathrm{pass}@k$ across six benchmarks as the candidate budget grows for
architectural and temperature sampling.}
\label{fig:qwen35_4b_scaling}
\end{figure}

Figures~\ref{fig:qwen3_8b_scaling}, \ref{fig:qwen25_7b_scaling}, and
\ref{fig:qwen35_4b_scaling} trace $\mathrm{pass}@k$ as candidates are added to
three representative checkpoints, extending Table~\ref{tab:main} beyond the
fixed budget of nine.

Across most plotted checkpoint--benchmark pairs, temperature sampling begins to
plateau earlier, while architectural sampling continues to gain as candidates
are added. Because the pattern is not uniform across panels, we treat these
curves as descriptive views of the candidate pools.

\section{Computation Cost and Latency}
\label{sec:cost}

Architectural sampling leaves the parameter count unchanged but traverses some
decoder layers more than once. We count these layer applications as an analytic
proxy before reporting latency for an $8\mathrm{B}$-parameter model on one
NVIDIA A100.

\paragraph{Cost model.}
\label{sec:cost-model}

Let a decoder have $L$ layers and a configuration reuse a contiguous window $B$
of size $|B|=b-a+1$ at depth $r$. Treating each decoder-layer application
equally, the configuration evaluates the prefix once, the window $r$ times,
and the suffix once:
\[
  N_{\text{layers}}(B,r)
  =\underbrace{(a-1)}_{\text{prefix}}
  +\underbrace{r\,|B|}_{\text{reused window}}
  +\underbrace{(L-b)}_{\text{suffix}}
  = L + (r-1)\,|B|.
\]
Normalizing one standard decoder traversal ($r=1$) to unit cost gives
\[
  \boxed{\;
  C(B,r)=\frac{N_{\text{layers}}(B,r)}{L}=1+\frac{(r-1)\,|B|}{L}.
  \;}
\]
Each additional traversal adds $|B|$ layer applications. The overhead grows
with window size and recursion depth and forms a smaller fraction of a deeper
decoder. Table~\ref{tab:cost-grid} instantiates $C(B,r)$ for $L=32$.
\begin{table}[!ht]
\centering
\setlength{\tabcolsep}{15pt}
\renewcommand{\arraystretch}{1.0}
\footnotesize
\caption{\captionheading{Relative Layer-Application Cost}
For a decoder with $L=32$, normalized to one standard decoder traversal. Cost
grows linearly with window size $|B|$ and recursion depth $r$; the experimental
setting $|B|=3$ is highlighted.}
\label{tab:cost-grid}
\resizebox{0.55\linewidth}{!}{%
\begin{tabular}{@{}lccc@{}}
\toprule
\rowcolor{HeaderNavy}
\textbf{\textcolor{white}{Window size $|B|$}} &
\textbf{\textcolor{white}{$r=1$}} & \textbf{\textcolor{white}{$r=2$}} &
\textbf{\textcolor{white}{$r=4$}} \\
\midrule
\rowcolor{Band7Bours!18}
$2$ & $1.00\times$ & $1.06\times$ & $1.19\times$ \\
\rowcolor{Band7Bours}
$3$ & $1.00\times$ & $1.09\times$ & $1.28\times$ \\
\rowcolor{Band7Bours!18}
$4$ & $1.00\times$ & $1.13\times$ & $1.38\times$ \\
\rowcolor{Band7Bours!18}
$5$ & $1.00\times$ & $1.16\times$ & $1.47\times$ \\
\bottomrule
\end{tabular}%
}
\end{table}

\paragraph{Cost of the full candidate set.}
Our grid uses $|B|=3$ for all windows, $|\mathcal{B}|=4$, and
$\mathcal{R}=\{2,4\}$, along with one standard path at $r=1$. This gives $k=9$
candidates. The total decoder-layer cost, normalized to one standard traversal,
is
\[
  \underbrace{1}_{r=1}
  +\underbrace{4\bigl(1+\tfrac{3}{32}\bigr)}_{r=2}
  +\underbrace{4\bigl(1+\tfrac{9}{32}\bigr)}_{r=4}
  = 1 + 4.375 + 5.125 = 10.5 .
\]
Under this proxy, the nine architectural candidates require $10.5$ standard
decoder traversals, versus $9.0$ for nine temperature samples.

\paragraph{Latency.}
\label{sec:latency}

Table~\ref{tab:latency} reports the corresponding latency figures. The architectural pool takes $53.8$~s, versus $46.1$~s for nine temperature
samples; its $r=2$ and $r=4$ paths add $0.48$~s and $1.44$~s per candidate.

\begin{table}[H]
\centering
\setlength{\tabcolsep}{10pt}
\renewcommand{\arraystretch}{1.0}
\footnotesize
\caption{\captionheading{Wall-Clock Latency}
Per-candidate and total latencies for Qwen3-VL-8B using architectural sampling
versus an equal-sized temperature-sampling baseline. The full $9$-candidate
architectural set is highlighted.}
\label{tab:latency}
\resizebox{0.85\linewidth}{!}{%
\begin{tabular}{@{}lcccc@{}}
\toprule
\rowcolor{HeaderNavy}
\textbf{\textcolor{white}{Configuration}} &
\textbf{\textcolor{white}{\#\,cand.}} & \textbf{\textcolor{white}{Cost\,($\times$)}} &
\textbf{\textcolor{white}{Latency\,(s)}} & \textbf{\textcolor{white}{$\Delta$ vs.\ std.\,(s)}} \\
\midrule
\rowcolor{AvgCell}
Standard pass $(r{=}1)$ & $1$ & $1.00$ & $5.12$ & $+0.00$ \\
\rowcolor{Band7Bours!30}
Window $|B|{=}3,\ r{=}2$ & $4$ & $1.09$ & $5.60$ & $+0.48$ \\
\rowcolor{Band7Bours!55}
Window $|B|{=}3,\ r{=}4$ & $4$ & $1.28$ & $6.56$ & $+1.44$ \\
\midrule
\rowcolor{Band7Bours}
Arch.\ sampling (total) & $9$ & $10.5$ & $53.8$ & $--$ \\
\rowcolor{AvgCell}
Temp.\ sampling (total) & $9$ & $9.0$ & $46.1$ & $--$ \\
\bottomrule
\end{tabular}%
}
\end{table}

\FloatBarrier

%% file: tables/variance.tex
\begin{table}[!h]
\centering
\setlength{\tabcolsep}{10pt}
\renewcommand{\arraystretch}{1.0}
\footnotesize
\caption{\captionheading{Run-to-Run Variation}
Mean $\pm$ standard deviation of $\mathrm{pass}@9$; the best setting per
benchmark is bold.}
\label{tab:variance}
\resizebox{0.85\linewidth}{!}{%
\begin{tabular}{@{}lccccc@{}}
\toprule
\rowcolor{HeaderNavy}
 &
\textbf{\textcolor{white}{CV-Bench}} & \textbf{\textcolor{white}{MMStar}} &
\textbf{\textcolor{white}{AI2D}} & \textbf{\textcolor{white}{A-OKVQA}} &
\textbf{\textcolor{white}{ScienceQA}} \\
\midrule
\rowcolor{AvgCell}
Temperature &
$88.48\,{\scriptstyle\pm 0.11}$ & $58.93\,{\scriptstyle\pm 0.14}$ &
$84.89\,{\scriptstyle\pm 0.23}$ & $88.65\,{\scriptstyle\pm 0.11}$ &
$88.75\,{\scriptstyle\pm 0.09}$ \\
\midrule
\rowcolor{Band7Bours}
Architectural &
$\best{96.40}\,{\scriptstyle\pm 0.18}$ & $\best{82.73}\,{\scriptstyle\pm 0.12}$ &
$\best{93.48}\,{\scriptstyle\pm 0.31}$ & $\best{95.72}\,{\scriptstyle\pm 0.19}$ &
$\best{97.97}\,{\scriptstyle\pm 0.25}$ \\
\bottomrule
\end{tabular}%
}
\end{table}

%% file: tables/model_families.tex
\begin{table}[H]
\centering
\setlength{\tabcolsep}{10pt}
\renewcommand{\arraystretch}{1.1}
\footnotesize
\caption{\captionheading{Generalization Beyond Qwen}
$\mathrm{pass}@9$ on InternVL3-2B and Gemma3-4B. $\Delta$ denotes architectural
minus temperature sampling; the best score per benchmark is bold.}
\label{tab:generalization-family}
\resizebox{0.85\linewidth}{!}{%
\begin{tabular}{@{}lcccccc@{}}
\toprule
\rowcolor{HeaderNavy}
 &
\textbf{\textcolor{white}{CV-Bench}} & \textbf{\textcolor{white}{MMStar}} &
\textbf{\textcolor{white}{AI2D}} & \textbf{\textcolor{white}{A-OKVQA}} &
\textbf{\textcolor{white}{ScienceQA}} & \textbf{\textcolor{white}{Avg.}} \\
\midrule

\rowcolor{Group7B}
\multicolumn{6}{@{}l}{\textbf{\;InternVL3-2B}} & \cellcolor{AvgCell} \\
\rowcolor{Band7B}
Temperature & 94.62 & 85.07 & \best{93.55} & 94.50 & 96.28 & \avg{92.80} \\
\rowcolor{Band7Bours}
Architectural & \best{95.87} & \best{87.87} & 93.52 & \best{96.16} & \best{98.31} & \avg{\best{94.35}} \\
\rowcolor{Band7B}
$\Delta$ & \gain{1.25} & \gain{2.80} & \drop{0.03} & \gain{1.66} & \gain{2.03} & \avg{\gain{1.54}} \\
\midrule
\rowcolor{Group3B}
\multicolumn{6}{@{}l}{\textbf{\;Gemma3-4B}} & \cellcolor{AvgCell} \\
\rowcolor{Band3B}
Temperature & 57.96 & 45.73 & 72.19 & 80.00 & 74.67 & \avg{66.11} \\
\rowcolor{Band3Bours}
Architectural & \best{81.27} & \best{70.13} & \best{87.26} & \best{91.88} & \best{88.15} & \avg{\best{83.74}} \\
\rowcolor{Band3B}
$\Delta$ & \gain{23.31} & \gain{24.40} & \gain{15.07} & \gain{11.88} & \gain{13.48} & \avg{\gain{17.63}} \\
\bottomrule
\end{tabular}%
}
\end{table}